\documentclass[times,final,5p]{elsarticle}

\usepackage{lineno,hyperref}
\usepackage{graphicx}
\usepackage{amsmath}
\usepackage{amssymb}
\usepackage{booktabs}
\usepackage{multirow}
\usepackage{threeparttable}
\usepackage{color}
\modulolinenumbers[5]

\journal{Pattern Recognition}

\begin{document}
	\begin{frontmatter}


		%
		

		
		
		
		
		
		
		
		
		\bibliographystyle{elsarticle-num}

		\title{LLHA-Net: A Hierarchical Attention Network for Two-View Correspondence Learning}
		
		\author[label1]{Shuyuan Lin}
		\ead{swin.shuyuan.lin@gmail.com}
		
		\author[label2]{Yu Guo}
		
		\author[label1]{Xiao Chen}
		
		\author[label3]{Yanjie Liang}
		
		\author[label4]{Guobao Xiao}

		\author[label1]{Feiran Huang \corref{cor1}}
		

		\cortext[cor1]{Corresponding author.}
		
		\address[label1]{College of Cyber Security, Jinan University, Guangzhou 510632, China.}
		\address[label2]{School of computer Engineering / School of Data Science, Guangzhou City University of Technology, 510800, China.}
		\address[label3]{Department of Strategic and Advanced Interdisciplinary Research, Peng Cheng Laboratory, Shenzhen 518000, China}
		\address[label4]{School of Electronics and Information Engineering, Tongji University, Shanghai, 201804, China}
		
		\begin{abstract}
			Establishing the correct correspondence of feature points is a fundamental task in computer vision. However, the presence of numerous outliers among the feature points can significantly affect the matching results, reducing the accuracy and robustness of the process. Furthermore, a challenge arises when dealing with a large proportion of outliers: how to ensure the extraction of high-quality information while reducing errors caused by negative samples. To address these issues, in this paper, we propose a novel method called Layer-by-Layer Hierarchical Attention Network, which enhances the precision of feature point matching in computer vision by addressing the issue of outliers. Our method incorporates stage fusion, hierarchical extraction, and an attention mechanism to improve the network's representation capability by emphasizing the rich semantic information of feature points. Specifically, we introduce a layer-by-layer channel fusion module, which preserves the feature semantic information from each stage and achieves overall fusion, thereby enhancing the representation capability of the feature points. Additionally, we design a hierarchical attention module that adaptively captures and fuses global perception and structural semantic information using an attention mechanism. Finally, we propose two architectures to extract and integrate features, thereby improving the adaptability of our network. We conduct experiments on two public datasets, namely YFCC100M and SUN3D, and the results demonstrate that our proposed method outperforms several state-of-the-art techniques in both outlier removal and camera pose estimation. Source code is available at http://www.linshuyuan.com.
		\end{abstract}

		
	\end{frontmatter}
	

	\section{Introduction}
	\label{sec:intro}
	
	Two-view matching is a fundamental task in computer vision \cite{shi2023jra} that aims to establish correspondences between feature points in images captured from two different viewpoints. It is closely related to many other computer vision tasks, such as structure from motion \cite{lin2023multimotion}, image retrieval \cite{lines}, 3D reconstruction \cite{xiao2024latent}, and remote sensing images \cite{chen2024ssl}. 
	The process of two-view matching involves extracting feature points from two different perspective images of the same scene and then matching these feature points to determine the relative position and pose between the two images. 
	The presence of numerous outliers can significantly undermine the reliability of image matching applications. Therefore, accurately extracting high-quality feature point information is crucial for removing outliers.
	However, this task is challenging due to various factors that can affect the accuracy of the results, such as rotation, translation, illumination changes, environmental factors, blur, and occlusion.
	To address these challenges, various techniques \cite{lin2024robust} have been developed, including feature point detection and description, feature matching, outlier removal, and geometric verification. 
	Several methods have focused on addressing key challenges such as outlier rejection, feature matching, and robustness to noise. For instance, MSA-Net \cite{MSA-net} introduced a multi-scale feature fusion strategy to improve feature matching accuracy under varying conditions, while T-Net \cite{T-Net} utilized attention mechanisms to enhance feature relevance. However, these methods still face limitations when applied to complex, noisy datasets, particularly in distinguishing relevant features from outliers.
	The method we designed aims to extract and fuse information for the feature points in these harsh environments to achieve good performance in outlier removal.
	In recent years, deep learning-based methods have also shown their great potentials in two-view matching, as they can learn more robust and discriminative features from large-scale data. 
	Learning-based methods such as OA-Net\cite{oanet} outperform traditional methods such as RANSAC\cite{RANSAC} across various experimental metrics.
	Despite the challenges, two-view matching remains an active research area, and further advancements in this field are expected to enhance the capabilities of computer vision systems in various domains.
	In this work, we hypothesize that a hierarchical attention mechanism combined with layer-by-layer feature fusion can significantly improve the accuracy and robustness of outlier removal in two-view correspondence learning, especially in scenarios with high outlier ratios and complex noise patterns.
	
	\begin{figure*}[!t]
		\centering
		-cleanVersion	\includegraphics[width=1\linewidth]{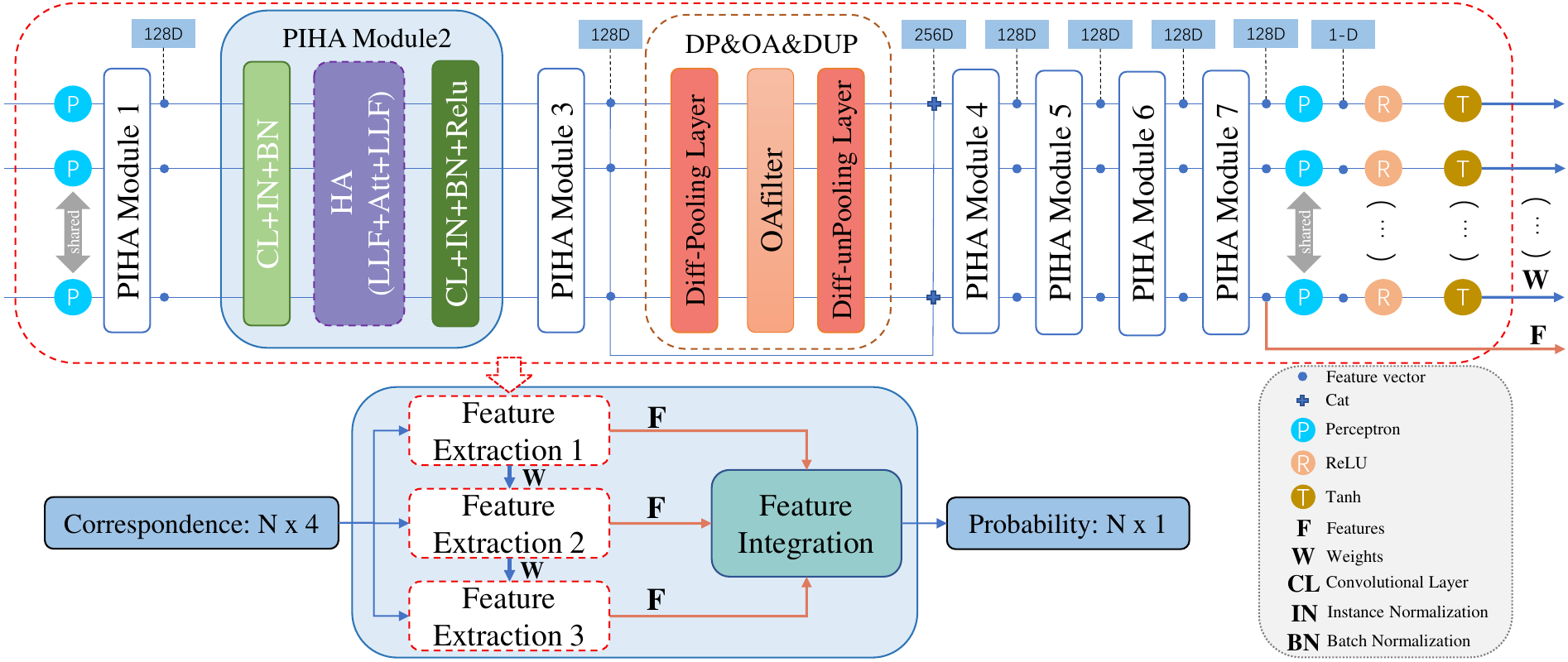}
		\caption{LLHA-Net: Overall network architecture and feature integration modules. The upper part illustrates the feature extraction architecture, while the lower part shows the overall network. The feature integration architecture includes multiple PIHA modules.}
		\label{fig:LLHA-Net}
	\end{figure*}\par

	Given two pictures, extracting the feature points and establishing a point-to-point correspondence is not a very difficult problem.
	However, matching feature points often generates a large number of outlier matching pairs, and it is challenging to find the inlier correspondences from a large number of mismatches, especially when the proportion of outlier correspondences is as high as $90\%$ \cite{linco}.
	HRMP \cite{linHRMP} reduces sensitivity to outliers through message propagation, while HOMF applies spectral clustering and hyperedge optimization to adaptively fit inliers.
	The outlier removal correspondence learning method includes the following three steps: obtaining feature points and their descriptors; establishing an initial correspondence set; and removing incorrect correspondence sets. Generally, the first step is achieved using existing methods, such as SIFT \cite{SIFT} and SuperPoint \cite{SUper-Point}. Then, the nearest neighbor method is used to build the initial set of correspondences. However, this process often produces a large number of outliers, far overwhelming the inliers. Therefore, the network for removing outliers becomes particularly important.\par

	To address the aforementioned issues, we propose a new and efficient neural network model called the layer-by-layer hierarchical attention network (LLHA-Net), as illustrated in Figure \ref{fig:LLHA-Net}. 
	LLHA-Net fully exploits an architecture with an iterative structure for information extraction to prevent being misled by a large amount of outlier information. 
	Specifically, we propose a novel layer-by-layer channel fusion (LLF) module to effectively fuse information. 
	This module generates and synthesizes information step by step within a multi-layer network, ensuring the richness of feature information as much as possible.
	We design a hierarchical attention (HA) module to extract feature information at various levels and fuse them using an attention mechanism. To ensure the permutation invariance of the module, a convolutional layer with 1 × 1 convolution kernel is added to the HA module, resulting in the permutation invariance hierarchical attention (PIHA) module. The PIHA module can extract feature information from coarse-to-fine levels and integrate them to maximize the internal point information and reduce the presence of outlier information.
	By employing multiple iterations, our framework further screens the original and redundant information, gradually improving the network's performance and enhancing the adaptability and generalization of the network model. To integrate the amount of information from multiple stages, we propose an information integration module that combines the information from each stage to obtain a more accurate and robust feature representation.
	Unlike methods that focus solely on low-level feature fusion or single-scale attention, LLHA-Net introduces a novel hierarchical attention mechanism that facilitates multi-level feature fusion, resulting in a more refined and accurate matching process. Moreover, the hierarchical attention mechanism enhances both the feature fusion process and the overall robustness of the model, ensuring more accurate correspondences even in the presence of significant noise.
	In this way, LLHA-Net can effectively remove the interference of outlier information, enhance the accuracy and stability of the model, and perform better in various application scenarios. Although the proposed LLHA-Net significantly improves the accuracy and robustness of feature point matching, it still presents certain limitations. In particular, in the iterations at each stage, the parameters and running time of the model have also been increased. 
	An overview of the proposed framework is shown in Fig.~\ref{fig:LLHA-Net}.
	The main contributions of the proposed method can be summarized as follows:
	\begin{itemize}
		\item We propose the layer-by-layer hierarchical attention network (LLHA-Net), which adopts an iterative structure information extraction architecture and ensures the richness of feature information in a multi-layer network by gradually generating and synthesizing information. This approach is effective to avoid the interference of abnormal information and enhances the network's adaptability and generalization.
		\item We propose the layer-by-layer channel fusion (LLF) module, which can preserve the key information at each stage while achieving overall fusion. This module can effectively fuses channel information. 
		It enhances the representation capability of feature points and improves the accuracy of feature point matching.
		\item We design the permutation invariance hierarchical attention (PIHA) module, which handles sparse correspondences in a permutation invariance manner while capturing both global perception and structural semantic information. Additionally, it performs information fusion at different levels to obtain more comprehensive and layered feature information.
	\end{itemize}\par
	The rest of this paper is structured as follows. 
	In Section \ref{related work}, we provide a review of the related work in the field. 
	Section \ref{proposed} presents a detailed description of our proposed method. 
	We discuss the experimental results and conduct ablation studies in Section \ref{experiment}.
	Finally, in Section \ref{conclusion}, we draw conclusions based on our findings.

	\section{Related Work}
	\label{related work}
	In this section, we will provide a brief introduction to traditional methods and learning-based methods for two-view feature point matching, followed by an overview of the attention mechanism and its development.
	\subsection{Traditional Methods}
	The traditional methods for removing outlier correspondences primarily include RANSAC \cite{RANSAC} and its variants \cite{GC-RANSAC, MLESAC}. These methods share a common approach of using hypothesis and verification techniques to identify the most likely set of correspondences. For instance, MLESAC \cite{MLESAC} is a random sampling consensus algorithm based on maximum likelihood estimation, which addresses the problem of mismatches in model fitting. GC-RANSAC \cite{GC-RANSAC} employs a graph cuts-based method to separate data points into inliers and outliers, reducing the impact of random sampling in the RANSAC algorithm and computing model parameters accordingly. PROSAC \cite{PROSAC} enhances the representativeness of sampled data by dynamically adjusting the sampling probabilities, thereby improving the accuracy and robustness of model estimation. 
	However, these methods also share some common drawbacks. Firstly, they tend to have low efficiency, often requiring a large number of iterations to find the optimal model parameters, resulting in lengthy computation times. Secondly, the results of the algorithms can be influenced by random sampling, necessitating multiple random samples to obtain stable outcomes. Thirdly, the algorithms may produce estimation errors when dealing with noisy datasets, leading to  inaccurate matching results.
	\subsection{Learning-Based Methods}
	With the advancement of deep learning and computational resources, numerous end-to-end methods \cite{lin2025mgcanet} based on neural networks have emerged for feature point matching, capable of generating pixel-level correspondences. SuperGlue \cite{superglue} employs attention mechanisms and graph neural networks to generate descriptors for feature points, combining local and global descriptors to calculate the probability of correct matching and obtain feature point correspondences. LoFTR \cite{LOFTR} utilizes a Transformer for detector-free feature matching, establishing rough matching using the attention mechanism and refining it by adjusting the receptive field. The global receptive field provided by the Transformer enables dense pixel-level matching, even in low-texture areas. COTR \cite{COTR} proposes a functional communication architecture that combines the advantages of dense and sparse methods, utilizing Transformer to capture global and local semantic information, achieving high-precision feature correspondence.
	MSGSA \cite{lin2024multistage} introduces a multi-stage hierarchical strategy that improves the robustness of correspondence learning, leveraging progressive refinement across multiple levels to address the challenges of large-scale outliers and complex geometric transformations.
	
	In another direction, some works \cite{oanet,T-Net} achieve matching by classifying feature points in the image as inliers or outliers. The steps of learning-based false matching removal methods typically involve: obtaining feature points and their descriptors, establishing an initial correspondence set, and then removing incorrect correspondence sets. Specifically, feature points are generated using methods like SIFT \cite{SIFT} and SuperPoint \cite{SUper-Point}, and feature point correspondences are established through nearest neighbor matching. 
	However, this process often generates a considerable proportion of feature point mismatches, making precise mismatch removal crucial. 
	PointNet \cite{qi2017pointnet} represents input data as point coordinates and ensures permutation invariance through maximum pooling. However, it fails to capture the local structure introduced by metric space points, limiting its ability to recognize fine-grained patterns and generalize to complex scenes. PointNet++ \cite{qi2017pointnet++} introduces multi-scale feature extraction and a hierarchical neural network to recursively divide the input point set. Nonetheless, the vector calculation process of feature points lacks consideration of point relationships. Acne \cite{sun2020acne} integrates the PointNet \cite{qi2017pointnet} architecture and context normalization for classifying feature point correspondences, focusing on global features but overlooking point similarity. OA-Net \cite{oanet} divides points into clusters, extracting local feature information within each cluster and distributing it to the original nodes.
	Progressive correspondence pruning \cite{progressive} employs a local-to-global consensus learning approach to progressively trim correspondences, using a ``pruning" module to discern reliable candidates from initial matches via consensus scores from dynamic graphs, and stacking pruning blocks sequentially for continuous refinement.
	LMCNet \cite{liu2021learnable} learns the motion coherence property for correspondence pruning.
	T-Net \cite{T-Net} introduces a feature extraction structure and a feature integration structure, iteratively synthesizing them. 
	However, the permutation-equivariant context squeeze-and-excitation (PCSE) module cannot make full use of the feature information by only using the squeeze-extraction strategy in the feature extraction process.
	PESA \cite{zhong2022pesa} obtains feature information from multiple perspectives through iteration, but each iteration solely relies on the results of the previous iteration, potentially leading to information loss. MSA-Net \cite{MSA-net} fully exploits a multi-scale attention module to enhance robustness to mismatched points, improving the representation ability of feature maps and extracting information context with fewer channel and spatial parameters. However, the context channel refine block can only employ information fusion from the previous stage and fails to retain information from each stage.
	NCMNet \cite{NCMnet} introduces a global-graph space, which explicitly captures long-range dependencies among correspondences, to seek consistent neighbors.
	PGFNet \cite{pfgnet} designs an iterative filtering structure to learn the preference scores of correspondences, guiding the correspondence filtering strategy.
	
	\par
	In contrast, our focus is on extracting and fusing semantic information by leveraging the information from each stage of the multi-stage architecture and adapting the fusion of information at different scales and levels.

	
	\subsection{Attention Mechanism}
	
	The attention mechanism is a weighted processing method employed in neural networks to flexibly process different parts of the input. Initially introduced in natural language processing \cite{yang2016hierarchical, attentionisallyouneed}, the attention mechanism has found widespread application in other fields such as computer vision and recommendation systems \cite{DualAttentionNetworkforSceneSegmentation, ma2022correspondencea} .
	Seq2Seq \cite{seq2seq} improves the relevance of the output sequence by assigning different weights during the decoding process using the attention mechanism. HAN \cite{yang2016hierarchical} proposes a text classification model based on hierarchical attention. It captures important information from the input text at various levels, thereby enhancing the accuracy of text classification. Transformer \cite{attentionisallyouneed} introduces a self-attention mechanism to enable the model to capture global information in the input sequence, resulting in improved accuracy.
	ViT \cite{VIT} introduces a cross-attention mechanism that divides an image into multiple patches for better handling relationships between various parts in images. DANet \cite{DualAttentionNetworkforSceneSegmentation} proposes dual attention networks, capable of simultaneously attending to both spatial and channel information in the input image. STN \cite{jaderberg2015spatial} proposes an attention-based spatial transformer. The network learns a transformation matrix for operations such as translation, rotation, and scaling on the input image, adaptively adjusting the transformation matrix based on the image content.
	CAT \cite{ma2022correspondencea} designs an attention-style structure to aggregate features from all correspondences and proposes a covariance-normalized channel attention to reduce the memory consumption and parameter scale.
	
	These methods, leveraging attention mechanisms, have made significant progress by focusing on the most relevant features for correspondence matching. However, they generally operate at a single scale, which restricts their ability to adapt to different levels of feature abstraction. In contrast, LLHA-Net incorporates a hierarchical attention mechanism that functions across multiple levels of abstraction. This design allows for more precise control over the feature fusion process, enhancing the model's capability to manage both global and local feature interactions. Furthermore, LLHA-Net's approach effectively mitigates the impact of outliers by concentrating on the most relevant features at each level.

	In this work, we enhance our network with an adaptive approach leveraging the attention mechanism. Unlike conventional methods with fixed weights for integrating global perception and structural semantics, our model employs adaptive learning to dynamically adjust weights based on input relevance. This leads to more effective information synthesis and improved performance across diverse contexts.

	
	\section{Proposed Method}
	\label{proposed}
	In the field of two-view correspondence learning, both deep feature information and shallow feature information are very important. Deep features capture high-level abstractions of the data, while shallow features contain rich local details and texture information. In order to make full use of both types of information, our method focuses on information interaction at the channel level and employs an innovative layer-by-layer generation and layer-by-layer fusion mechanism.
	In this section, we present a comprehensive description of the LLHA-Net architecture. We start by outlining the problem formulation, followed by a detailed explanation of the layer-by-layer channel fusion module, the hierarchical attention module, the permutation-invariance hierarchical attention module, and the loss function employed in our network. Lastly, we provide a summary of the overall network architecture.
	
	our network architecture, including the layer-by-layer channel fusion module, the hierarchical attention module and the permutation invariance hierarchical attention module. Finally, we will introduce the loss function used in our network.

	\subsection{Problem Formulation}
	Two-view correspondence learning involves two tasks: an inlier/outlier classification task and an essential matrix regression task.
	Our objective is to establish accurate correspondences and recover the relative pose between a pair of images in the same scene, denoted as $(I, I')$. 
	To achieve this, we begin by detecting feature points and extracting descriptors using traditional methods such as SIFT or learning-based methods such as SuperPoint. 
	Next, we employ the nearest neighbor search to establish initial correspondences for the feature points. The process is as follows:\par
	\begin{equation}
		S=[s_{1} , s_{2} , s_{3}, \dots , s_{N}]\in \mathbb{R}^{N\times 4},
		\label{eq:1}
	\end{equation}
	\begin{equation}
		s_{i}=(x_{i} , y_{i} , x_{i}^{\prime} , y_{i}^{\prime}),
		\label{eq:2}
	\end{equation}
	where $S$ represents the input correspondences of the outlier removal method; $s_{i}$ represents a putative correspondence; $(x_{i}, y_{i})$ and $(x_{i}^{\prime}, y_{i}^{\prime})$ represent the coordinate position corresponding to the $i$-th group in the two images.\par
	In the outlier removal task, the final output is the probability set $P$, which represents the likelihood of correspondences being inliers, as in:
	\begin{equation}
		P=\left \{p_{1}, p_{2}, p_{3}, \dots , p_{N}\right \}, p_{i}\in [0, 1), 
		\label{eq:3}
	\end{equation}
	where $p_{i}$ represents the probability of $i$-th correspondence being an inlier. A value of $p_{i}=0$ indicates that the $i$-th correspondence is an outlier. 
	We employ the weighted eight-point algorithm to regress the essential matrix based on $P$.
	The overall structure can be written as follows:
	\begin{equation}
		P=H_{all} (S),
		\label{eq:4}
	\end{equation}
	\begin{equation}
		\hat{E} =g(P, S),
		\label{eq:5}
	\end{equation}
	where $H_{all}$ represents the whole outlier removal network; $\hat{E}$ represents the regressed essential matrix; and $g$ is the weighted eight-point algorithm used to compute the essential matrix $\hat{E}$.

	\subsection{Layer-by-Layer Channel Fusion Module}
	\begin{figure}[t]
		\centering
		\includegraphics[width=0.7\linewidth]{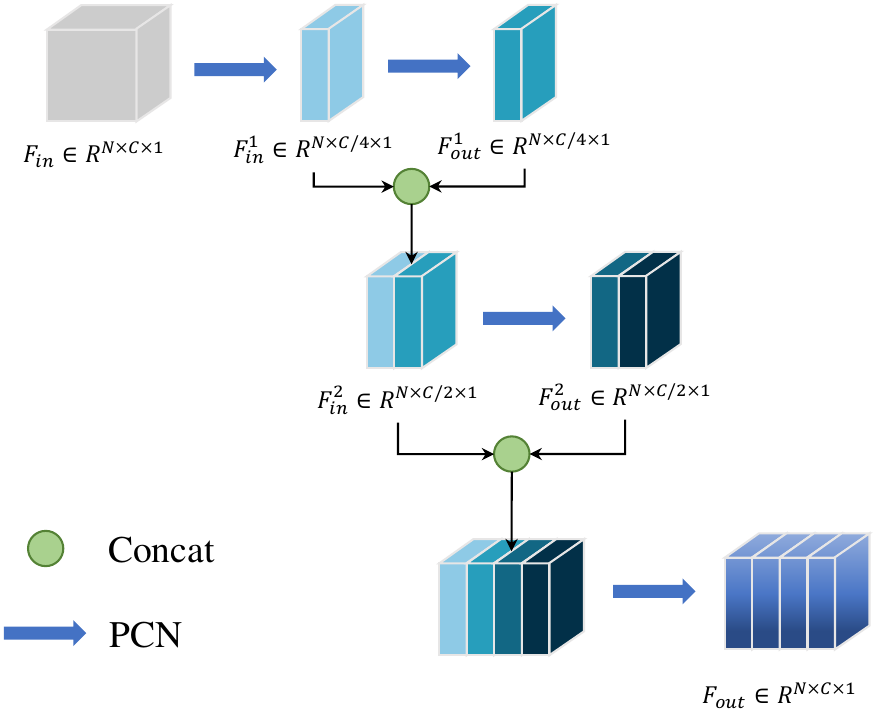}
		\caption{LLF module: layer-by-layer channel fusion module.}
		\label{fig:LLF}
	\end{figure}
	In a multi-layer network architecture, the final output is primarily determined by the data from the preceding stage, which may result in the loss of important semantic information. In addition, the context channel refinement block in MSA-Net \cite{MSA-net} seeks to enhance the processing at the final stage but has limited effectiveness in integrating information from earlier intermediate stages.
	To address this problem, we propose a novel LLF module, as shown in Figure \ref{fig:LLF}. The LLF module is designed to capture and preserve the feature semantic information at each stage layer-by-layer to achieve channel fusion. It enhances the representation capability of feature points and improves the accuracy of feature point matching.\par
	Given a feature map $F_{in} \in \mathbb{R}^{N \times C}$, we employ PointCN \cite{LFGC} with a weight of $C \times \frac{C}{h} \times 1 \times 1$ to downsample the feature map and remove redundant contextual information.
	
	Then, the channel extraction information of the first stage $F_{in}^{1}$ is generated as follows:
	\begin{equation}
		F_{in}^{1}  = PCN_{1} \left (F_{in} \right),
	\end{equation}
	where $PCN_{1}$ represents the PointCN in the first stage. 
	This approach allows for retaining the most important feature information while reducing computational complexity. To fuse the input and output information of the previous layer, we use the concatenation operation (Concat) to combine the two sets of information. As a result, the output information of the next layer has double the dimension of the output information of the previous layer.
	Next, we apply PointCN with the corresponding doubled dimension to fuse the feature information generated by the previous layer. This process preserves the semantic information of each stage and allows different $PCN_{i}, i\ in \left \{ 2, 3, ..., h\right \}$ to generate new feature information, enriching the information of our feature map, as follows:
	\begin{equation}
		F_{in}^{i} = Concat(F_{in}^{i-1}, F_{out}^{i-1}),
	\end{equation}
	\begin{equation}
		F_{out}^{i} = PCN_{i} \left (F_{in}^{i} \right),
	\end{equation}
	where $F_{in}^{i}$ represents the input of the $i$-th layer, and $F_{out}^{i}$ represents the output of the i-th layer.
	Therefore, by performing channel downsampling, we reduce redundant information in data and decrease the computational load. Subsequently, through layer-by-layer fusion and generation, the original dimension is restored, increasing the model's complexity, as well as enhancing its fitting ability and accuracy.\par
	In summary, the input for each stage is derived from both the input and output of the preceding stage. This design allows each stage to not only incorporate information from the previous stage but also integrate its own inherent characteristics. Consequently, the resulting output feature map exhibits enhanced representation capabilities and encapsulates more comprehensive semantic information.
	
	\subsection{Hierarchical Attention Module}
	\begin{figure*}[t]
		\centering
		\includegraphics[width=1\linewidth]{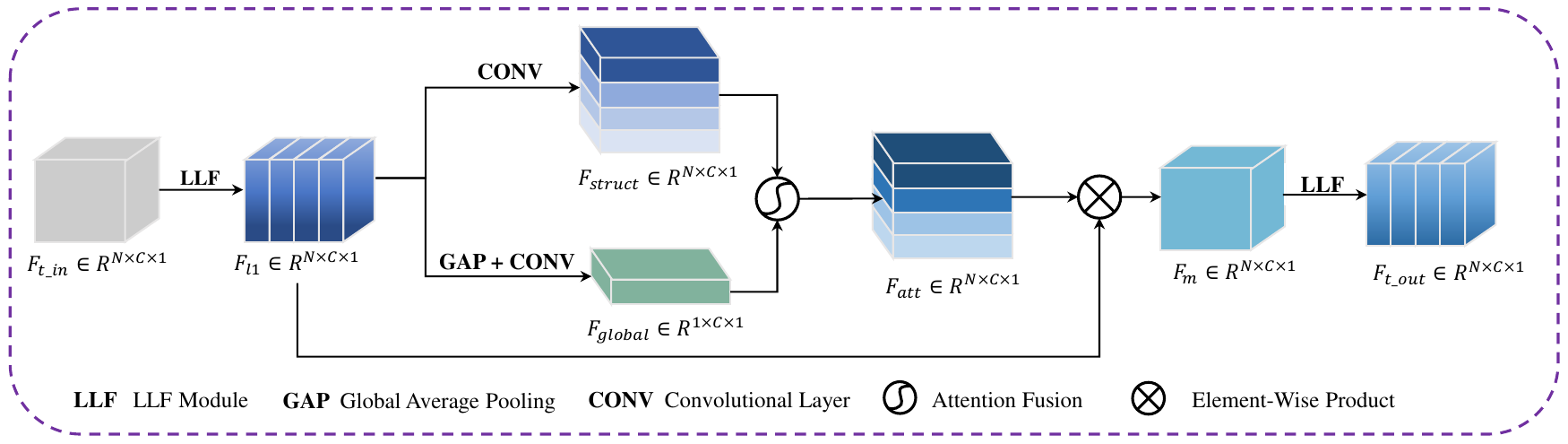}
		\caption{HA module: Hierarchical attention module uses the attention mechanism to adaptively fuse the information extracted from different hierarchical structures, while the LLF module fuses the information of each channel.}
		\label{fig:HA}
	\end{figure*}
	In real-world scenarios, several factors such as lighting conditions, image quality, and pixel variations can result in inlier correspondences accounting for only approximately 10\% of the total correspondences. 
	This poses a significant challenge for networks that rely on contextual normalization (CN) layers, as these layers treat all correspondences equally, which is not robust to outliers \cite{sun2020acne}. 
	Consequently, CN layers can significantly degrade network performance and compromise a substantial amount of semantic information.
	To overcome this issue, we develop a HA module that extracts diverse information blocks from features at different levels and incorporates an attention mechanism to adaptively fuse each information block. 
	This approach enables enhanced feature extraction and improved model performance.\par
\begin{figure}[!t]
		\centering
		\includegraphics[width=1\linewidth]{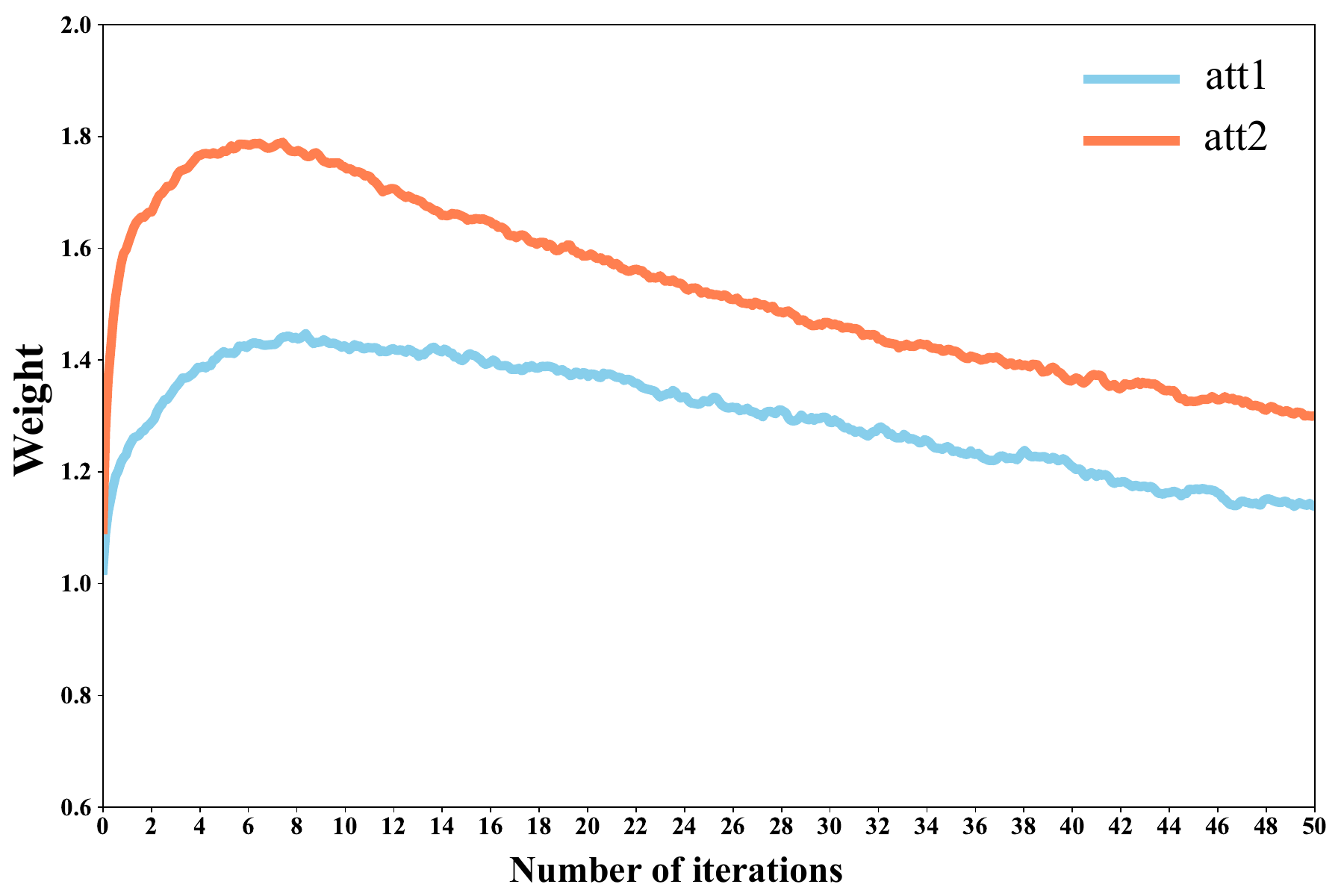}
		\caption{The weight variations of global perception information and local semantic information in the HA module are shown over 500,000 iterations. The horizontal axis represents the number of iterations, measured in units of ten thousand.}
		\label{att1andatt2}
	\end{figure}

	A crucial component of the HA module is the extraction of global perception and structural semantic information and their subsequent fusion.
	As depicted in Figure \ref{fig:HA}, 
	for a feature map $F_{t\_{in}}$, we first fuse its channel information through the LLF module, as follows:
	\begin{equation}
		F_{l1} = LLF(F_{t\_in}).
	\end{equation}
	Then, we utilize the global average pooling operation to extract global perception information from $F_{l1}$. 
	Global average pooling is a common feature extraction operation that reduces the dimensionality of the input feature map. 
	It compresses the feature map of each channel into a single value by taking the average of all values in the feature map, resulting in a global perception information representation $F_{global}$, as follows:
	\begin{equation}
		F_{global} = B(CONV(GAP(F_{l1})),
	\end{equation}
	where GAP represents the global average pooling operation, CONV represents a convolutional layer, and $B$ represents batch normalization. 
	On the other hand, the structural semantic information representation $F_{struct}$ is extracted only by convolution. This method can extract structural features and semantic information in different directions and scales, as in the following:
	\begin{equation}
		F_{struct} = B(CONV(F_{l1})),
	\end{equation}
	where $F_{struct}$ represents the structural semantic information. 
	Subsequently, we obtain weight factors for the subsequent attention mechanisms by attentively fusing the global perception and structural semantics information, as follows:
	\begin{equation}
		F_{att} = att_{1} \times F_{struct} + att_{2}\times F_{global},
	\end{equation}
	where $att_{1}$ and $att_{2}$ are adaptive parameters respectively, which are learned during the training process. 
	Their initial values are set to 1, which comes from the adaptive measurement of the importance of the two kinds of information.
	Figure \ref{att1andatt2} shows the variations in $att_{1}$ and $att_{2}$ over 500,000 iterations. The results indicate that $att_{2}$ consistently exceeds $att_{1}$, highlighting the dominant role of global perception information in the information interaction process.
	
	\begin{equation}
		F_{m} = F_{l1} \cdot F_{att},
	\end{equation}
	where $F_{att}$ represents the attention weight, $Att$ is the attention fusion function, and $F_{m}$ incorporates the global perception and structural semantics information of the feature map.\par
	To enhance the representation ability of features, it is important to not only fuse information at different scales but also perform channel information fusion on them. Therefore, we introduce the LLF module at the end of the HA module to obtain the output feature $F_{t\_out}$, as follows:
	\begin{equation}
		F_{t\_out} = LLF(F_{m}).
	\end{equation}\par
	The HA module not only fuses information across channels but also fuses information of different levels within a channel.
	This module allows our network to fully leverage the rich semantic information hidden between each feature point pair. 
	By combining the global perception and structural semantic information, the HA module enables the network to capture both the overall context and fine-grained details of feature point pairs between images.
	As a result, the network can extract more robust and discriminative features.
	\subsection{Permutation Invariance Hierarchical Attention Module}
	\label{PIHA} 
	In our task, the input consists of unordered pairs of coordinate points, with interior lines accounting for only 10\% of the total correspondences. Moreover, for the same pair of images, the extracted feature point coordinates and their corresponding matches can vary, and the input order of the coordinate point pairs is random. As a result, managing sparse correspondences and ensuring permutation invariance becomes a key challenge.
	To address the issue, we introduce the PIHA module, as depicted in Figure \ref{fig:LLHA-Net}. The PIHA module incorporates a convolutional layer block before and after the HA module, using a 1 × 1 convolutional kernel.
	The reason for choosing the $1\times1$ convolution kernel is that when the dimension of the convolution kernel exceeds 1, the sparse and disordered assumed correspondence is mixed in the form of a point cloud, thus destroying the permutation invariance of the model. 
	Additionally, Instance Normalization and Batch Normalization are incorporated to standardize the information, thereby expediting information convergence.
	This design guarantees permutation invariance, ensuring that each corresponding information can be embedded in the features regardless of the input order.\par
	In summary, the PIHA module ensures permutation invariance, enabling  our network to handle sparse correspondences in a permutation-invariant manner. Additionally, the module captures both global perception and structural semantics information while integrating information at various levels to obtain more comprehensive and layered feature information.
	
	\subsection{Loss Function}
	Following \cite{oanet, T-Net}, we employ a hybrid loss function to optimize our network:
	\begin{equation}
		\text { Loss }=l_{c}(P, L_{ab})+\alpha l_{mat}(\hat{E}, E),
	\end{equation}
	where $\text { Loss }$ is the mixture loss function of the whole network, $l_{c}(\cdot, \cdot)$ is a binary cross entropy loss between the probability set $P$ and the label set $L_{ab}$, which quantifies the discrepancy between $P$ and $L_{ab}$, serving as a metric for how well the model's predictions correspond to the true labels.  $l_{mat}(\cdot, \cdot)$ is the essential matrix loss between the predicted essential matrix $\hat{E}$ and the ground truth $E$, respectively. The essential matrix loss ensures that the network not only predicts accurate probabilities but also effectively captures the underlying geometric relationships. And $\alpha$ is the hyper-parameter we set to balance the weights of the classification loss and the essential matrix loss. Specifically, we calculate the essential matrix loss as follows MSA-Net \cite{MSA-net}:
	\begin{equation}
			l_{mat}(\hat{E}, E)=\frac{\left(v_{2}^\mathrm {T} \hat{E} v_{1}\right)^{2}}{(E v_{1})_{[1]}^{2}+(E v_{1})_{[2]}^{2}+\left(E^\mathrm {T} v_{2}\right)_{[1]}^{2}+\left(E^\mathrm {T} v_{2}\right)_{[2]}^{2}},
	\end{equation}
	where $v_{1}$ and $v_{2}$ represent two keypoint positions forming the correspondence.

	\subsection{Network Architecture}
	The architecture of LLHA-Net consists of two main parts, as illustrated in Figure \ref{fig:LLHA-Net}. The first part is the feature extraction architecture, which comprises three feature extraction modules. Each module consists of 7 PIHA modules and a DP\&OA\&DUP \cite{oanet} module. 
	The DP\&OA\&DUP module clusters the feature points to establish relationships among the clusters, enabling the network to model the global context more effectively. 
	The feature extraction modules generate weights and features at different stages, and these weights and initial correspondences are residually connected to the next stage. 
	Through three stages of iteration, feature outputs from three different stages are obtained.
	The second part is the feature integration architecture, which takes the three feature outputs obtained from the feature extraction architecture as input. 
	By synthesizing feature extraction information and performing interactive learning, more accurate and comprehensive features can be obtained.\par
	Overall, the feature extraction architecture synthesizes information at different depths to obtain more diverse and informative features. 
	The feature integration architecture performs interactive learning on the extracted features to improve the accuracy and robustness of the classification task. 
	LLHA-Net can obtain more accurate and comprehensive feature representations, getting better performance in feature matching and correspondence learning.

	\begin{table*}[!htbp]
		\centering
		\caption{Comparative results of outlier removal on the YFCC100M and SUN3D datasets. The font in \textbf{black} represents the highest effect in the same column.}
		\resizebox{0.95\linewidth}{!}{
			\begin{tabular}{ccccccccccccccccc}
				\toprule
				Datasets & \multicolumn{8}{c}{YFCC100M(\%)}                              & \multicolumn{8}{c}{SUN3D(\%)} \\
				\midrule
				\multirow{2}[4]{*}{Matcher} & \multicolumn{3}{c}{Known Scene} &       & \multicolumn{3}{c}{Unknown Scene} &       & \multicolumn{3}{c}{Known Scene} &       & \multicolumn{3}{c}{Unknown Scene} &  \\
				\cmidrule{2-4}\cmidrule{6-8}\cmidrule{10-12}\cmidrule{14-16}          & P(\%) & R(\%) & F(\%) &       & P(\%) & R(\%) & F(\%) &       & P(\%) & R(\%) & F(\%) &       & P(\%) & R(\%) & F(\%) &  \\
				\midrule
				RANSAC& 47.35 & 52.39 & 49.74 &       & 43.55 & 50.65 & 46.83 &       & 51.87 & 56.27 & 53.98 &       & 44.87 & 48.82 & 46.76 &  \\
				Point-Net++ & 49.62 & 86.19 & 62.98 &       & 46.39 & 84.17 & 59.81 &       & 52.89 & 86.25 & 65.57 &       & 46.30  & 82.72 & 59.37 &  \\
				CNe-Net& 54.43 & 86.88 & 66.93 &       & 52.84 & 85.68 & 65.37 &       & 53.70  & 87.03 & 66.42 &       & 46.11 & 83.92 & 59.52 &  \\
				DFE& 56.72 & 87.16 & 68.72 &       & 54.00    & 85.56 & 66.21 &       & 53.96 & 87.23 & 66.68 &       & 46.18 & 84.01 & 59.60  &  \\
				OA-Net& 58.90 & 88.87 & 70.85    &       & 55.77 & 86.34 & 67.77 &       & 54.44  & 88.64 & 67.45 &       & 46.32 & 84.46  & 59.83 &  \\
				T-Net& 62.18 & 89.26 & 73.30  &       & 58.20  & 86.01 & 69.42 &       & 55.35 & 88.43 & 68.08 &       & 47.17 & 84.05 & 60.43 &  \\
				PESA & 61.43 & 89.63 & 72.90  &       & 58.02 & 87.01 & 69.62 &       & 55.08 & 88.56 & 67.92 &       & 47.29 & 84.81 & \textbf{60.72} &  \\
				MSA-Net& 59.27 & \textbf{90.82} & 71.73 &       & 56.23 & \textbf{89.10}  & 68.95 &       & 53.45 & \textbf{88.65} & 66.69 &       & 45.72 & \textbf{85.02} & 59.46 &  \\
				\textbf{LLHA-Net}  & \textbf{63.24} & 90.14 & \textbf{74.33} &       & \textbf{59.28} & 87.34 & \textbf{70.62} &       & \textbf{55.53} & 88.51 & \textbf{68.24} &       & \textbf{47.37} & 84.04 & 60.59 &  \\
				\bottomrule
			\end{tabular}%
		}
		\label{tab:outlier removal}%
	\end{table*}%
	\begin{figure*}[!bp]
		\centering
		\includegraphics[width=1\linewidth]{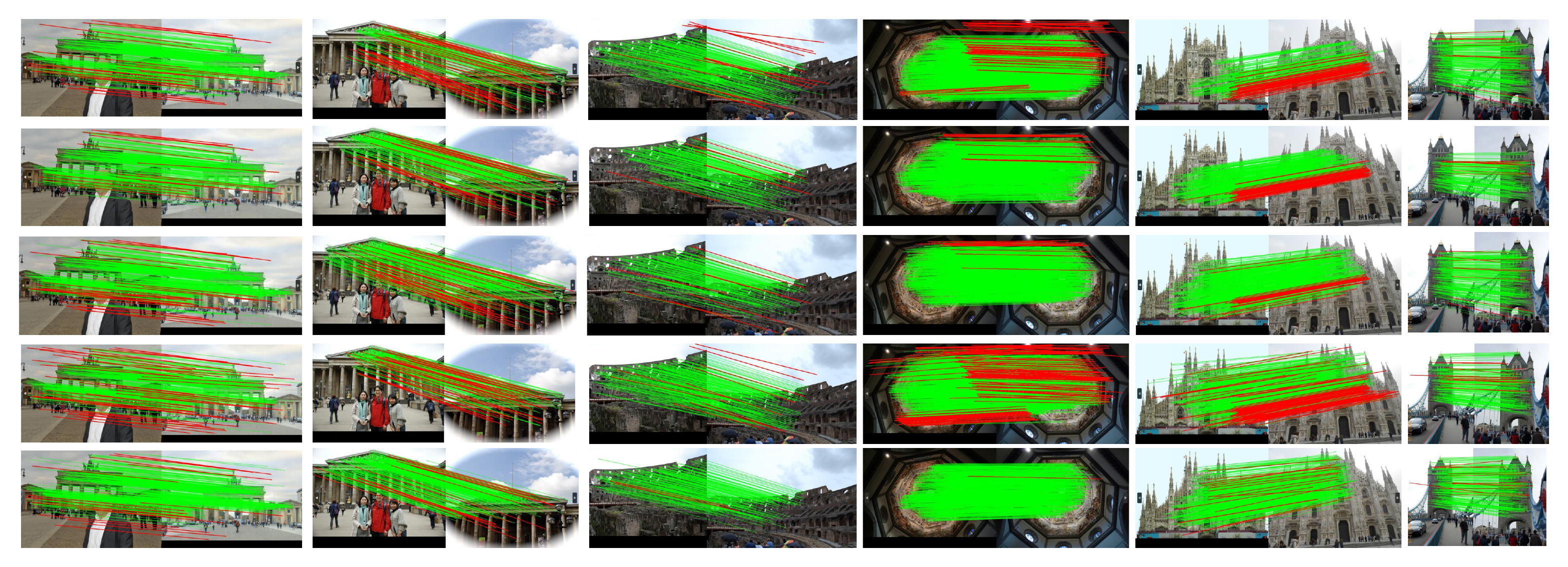}
		\caption{Some visualization results obtained using 5 competing methods on the YFCC100M dataset. Each row corresponds to a different method: the $1^{st}$ to $5^{th}$ rows represent OA-Net \cite{oanet}, T-Net \cite{T-Net}, PESA \cite{zhong2022pesa}, MSA-Net \cite{MSA-net}, and the proposed LLHA-Net, respectively. Red indicates remaining incorrect correspondences, while green indicates correct correspondences.}
		\label{fig:outlier}
	\end{figure*}
	
	\begin{figure*}[!htbp]
		\centering
		\includegraphics[width=1\linewidth]{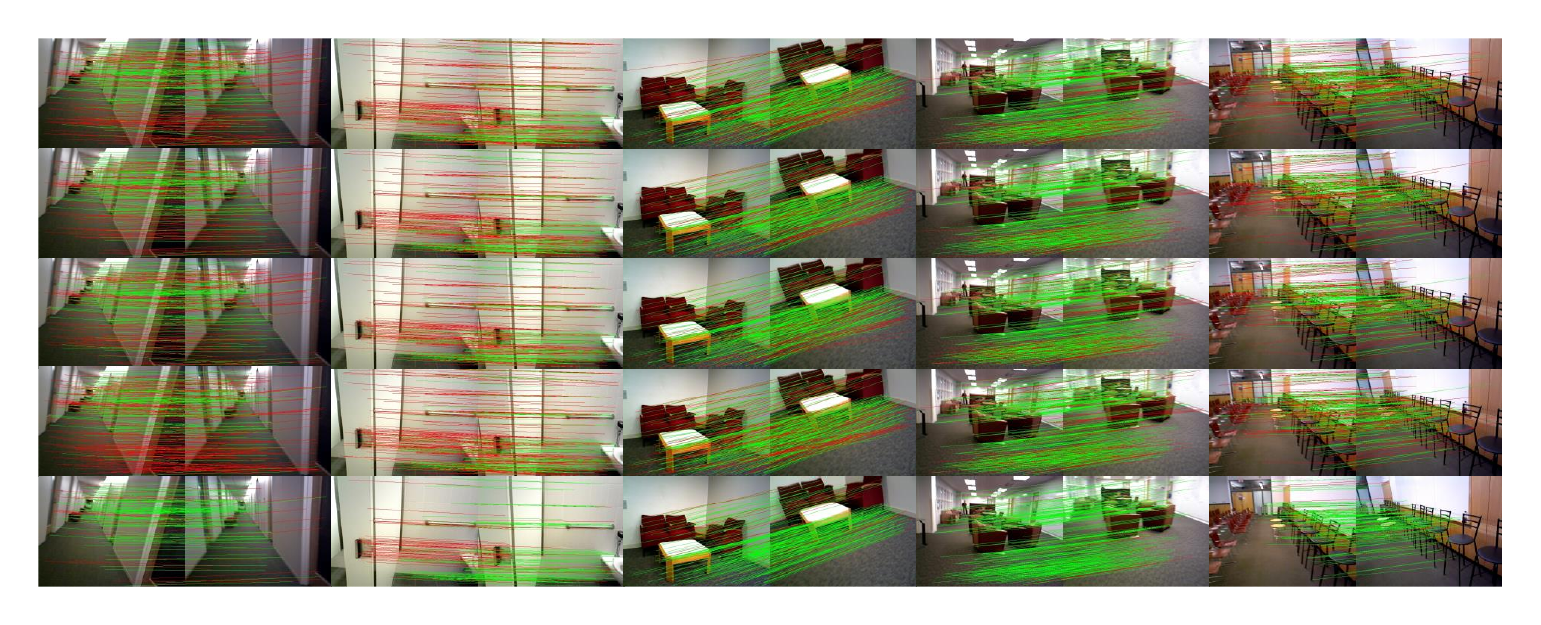}
		\caption{Some visualization results obtained using six competing methods on the SUN3D dataset are presented. Each row corresponds to a different method: the $1^{st}$ to $5^{th}$ rows represent OA-Net \cite{oanet}, T-Net \cite{T-Net}, PESA \cite{zhong2022pesa}, MSA-Net \cite{MSA-net}, and the proposed LLHA-Net, respectively. Red lines indicate remaining incorrect correspondences, while green lines represent correct correspondences.}
		\label{fig:sunoutlier}
	\end{figure*}

	\section{Experiment}
	\label{experiment}
	In this section, we provide an overview of the experiments conducted to assess the effectiveness of our proposed method. 
	First, we introduce the datasets utilized in the experiments and the evaluation metrics employed to measure the performance of our method. 
	Next, we delve into the experiments, covering aspects such as the implementation of our method and the configuration of hyperparameters. 
	Furthermore, we compare the performance of our method with existing methods in the tasks of outlier removal and camera pose estimation. Finally, we perform ablation experiments to assess the significance of different components within our method.
	
	\subsection{Datasets}
	The experimental datasets comprise both outdoor and indoor scenes. Each dataset is further divided into two categories: known scenes and unknown scenes. The unknown scenes are exclusively utilized as the test dataset. As for the known scenes, we split them into three parts: training ($60\%$), validation ($20\%$), and testing ($20\%$).

	\subsubsection{Outdoor Dataset}
	The YFCC100M dataset \cite{thomee2016yfcc100m}, introduced by Yahoo, is a vast collection of millions of images and videos encompassing various scenes and subjects. 
	It encompasses a wide range of content, including natural landscapes, people, animals, buildings, and more. 
	The dataset provides valuable features such as color histograms, SIFT features, HOG features, and others, which can be leveraged for image and video feature extraction and description. 
	Consequently, it has become an essential resource in the field of image and video analysis. 
	Following the approach in \cite{oanet}, we utilize four specific sequences, namely Buckingham Palace, Sacre Coeur, Reichstag, and Notre Dame front facade, as unknown scenes to assess the generalization capability of our method. 
	The remaining 68 sequences are employed as known scenes for training, validation, and testing purposes.
	
	\subsubsection{Indoor Dataset}
	The SUN3D dataset \cite{xiao2013sun3d} is a comprehensive indoor scene dataset that comprises diverse indoor environments such as offices, living rooms, restaurants, and more. 
	Each scene within the dataset is associated with multiple camera positions, providing RGB images, depth maps, and camera pose information. 
	The SUN3D dataset stands out due to its extensive data volume, rich scene variety, and detailed annotations. 
	Consistent with the approach outlined in \cite{T-Net}, we adopt the same dataset configuration for our experiments. 
	Specifically, we designate 15 scenes from the SUN3D dataset as unknown scenes, which are reserved exclusively for testing purposes. 
	The remaining 239 scenes are considered known scenes and are used for training and validation.
	
	It is worth noting that both the YFCC100M and SUN3D datasets contain challenging scenes, which are crucial for evaluating the robustness of competing methods and LLHA-Net. Specifically, the YFCC100M dataset features a diverse collection of images captured from various angles and distances, effectively simulating extreme viewpoint changes. For example, certain image sequences exhibit significant rotations and translations, posing challenges for accurate feature matching and pose estimation. Similarly, the SUN3D dataset includes indoor scenes with complex geometries, varying occlusions, and cluttered backgrounds. These characteristics often result in conditions with low inlier ratios, particularly in scenes where features are partially obscured or significant noise is present. By leveraging these challenging aspects of the datasets, we can evaluate the performance of competing methods and LLHA-Net under real-world conditions, demonstrating their capabilities in handling difficult scenarios.

	\subsection{Evaluation Metrics}
	As part of our evaluation process, we will focus on two key aspects: outlier removal and camera pose estimation. To assess the performance of our method in outlier removal, we employ three widely-used metrics: precision (P), recall (R), and F-score (F). These metrics evaluate how effectively our method identifies and removes outliers from the dataset. By considering P, R, and F together, we can comprehensively assess the quality of outlier removal.\par
	For camera pose estimation, we utilize the mean average precision (mAP) metric. 
	This metric takes both rotational and translational differences into account and it provides an overall evaluation of the angular disparity between the predicted and ground truth camera pose. 
	By employing mAP, we can assess the deviation in the camera pose relative to the objects present.
	
	\subsection{Implementation Details}
	We selected four classic networks (RANSAC \cite{RANSAC}, Point-Net++ \cite{qi2017pointnet}, CNe-Net \cite{LFGC}, and DFE \cite{DFE}) and four networks currently recognized for their high performance (OA-Net \cite{oanet}, T-Net \cite{T-Net}, PESA \cite{zhong2022pesa} and MSA-Net \cite{MSA-net}) as competing methods. 
	For the former, we referred to the experimental results in T-Net, and for the latter, we conducted experiments using the same dataset, experimental environment, and settings.
	It is worth noting that the same preprocessed data was used for all comparative experiments to ensure the consistency and reliability of the results.
	Our network is implemented on PyTorch, used a batch size of $32$ and the Adam solver with a learning rate of $10^{–4}$ to optimize the parameters. 
	The hyper-parameter $\alpha$ was set to $0$ for the first $2k$ iterations and $0.1$ for the remaining $480k$ iterations. 
	All experiments were conducted on NVIDIA 3090 GPUs.

	\subsection{Outlier Removal}
	
	Outlier removal is an important task in camera pose estimation, as it involves extracting key feature points from images and matching them to estimate the camera pose. However, due to the presence of a large number of mismatched feature points in the image, it is necessary to filter and exclude these outlier feature points to improve the accuracy of camera pose estimation.
	The results of our outlier removal experiments using the YFCC100M and SUN3D datasets, as well as results from other studies, are presented in Table \ref{tab:outlier removal}. 
	MSA-Net achieves the highest R, with LLHA-Net performing slightly lower than MSA-Net but better than all other networks. 
	The reason is that our network adopts an overly rich feature point information fusion method, resulting in some positive samples being misclassified as negative samples. 
	Nonetheless, our network achieves the best P and F. The P values are 63.24\%, 59.28\%, 55.53\% and 47.37\% in the four cases, which is a significant improvement, especially in YFCC100M scenes.
	This also indicates that our network extracts more accurate semantic and geometric information for feature correspondences.
	Similar to the other competing methods, LLHA-Net exhibits lower performance metrics on unknown scenes compared to known scenes. However, its performance on the unknown scenes closely aligns with that on the known scenes, demonstrating robustness relative to the other methods. Notably, the P and F-scores are near their peak values, highlighting LLHA-Net's generalization ability.
	Figures \ref{fig:outlier} and \ref{fig:sunoutlier} visualize the results of our method in some scenes of YFCC100M and SUN3D, showing that the outlier removal quality of LLHA-Net is significantly higher than other competing methods, addressing the common misclassification issues of existing methods. \par

	In summary, the experimental results demonstrate that our method outperforms other methods in outlier removal. LLHA-Net exhibits state-of-the-art performance in terms of P and F, and sub-optimal performance in terms of R. These results provide a guarantee of high-quality correspondences for the camera pose estimation task.
	
	\subsection{Camera Pose Estimation}
	\begin{table*}[!ht]
		\centering
		\caption{Performance comparison for camera pose estimation on YFCC100M and SUN3D datasets. The mAP performance at error thresholds 5° and 20° are reported \textbf{without/with} RANSAC post-processing.
			The font in \textbf{black} represents the highest effect in the same column.}
		\resizebox{0.95\linewidth}{!}{
			\begin{tabular}{ccccccccccccc}
				\toprule
				Datasets & \multicolumn{6}{c}{YFCC100M}  & \multicolumn{6}{c}{SUN3D} \\
				\midrule
				\multirow{2}[4]{*}{Matcher} & \multicolumn{2}{c}{Known Scene} &       & \multicolumn{2}{c}{Unknown Scene} &       & \multicolumn{2}{c}{Known Scene} &       & \multicolumn{2}{c}{Unknown Scene} &  \\
				\cmidrule{2-3}\cmidrule{5-6}\cmidrule{8-9}\cmidrule{11-12}          & 5°    & 20°   &       & 5°    & 20°   &       & 5°    & 20°   &       & 5°    & 20°   &  \\
				\midrule
				RANSAC & –/5.81 & –/16.88 &       & –/9.07 & –/22.92 &       & –/4.52 & –/15.46 &       & –/2.84 & –/11.19 &  \\
				Point-Net++& 10.49/33.78 & 31.17/56.24 &       & 16.48/46.25 & 42.09/67.53 &       & 10.58/19.17 & 35.75/44.06 &       & 8.10/15.29 & 30.97/35.83 &  \\
				CNe-Net& 13.81/34.55 & 35.20/57.27 &       & 23.95/48.03 & 52.44/69.10 &       & 11.55/20.60 & 36.12/44.33 &       & 9.30/16.40 & 31.32/37.23 &  \\
				DFE& 19.13/36.46 & 42.03/59.15 &       & 30.27/51.16 & 59.18/70.88 &       & 14.05/21.32 & 39.12/44.67 &       & 12.06/16.26 & 36.17/37.72 &  \\
				OA-Net & 32.90/42.32 & 56.76/64.39 &       & 41.73/53.20 & 68.84/73.31 &       & 21.45/22.78 & 48.57/47.11 &       & 15.69/17.11 & 41.02/39.16 &  \\
				T-Net& 40.74/45.07 & 63.66/66.80 &       & 44.95/55.75 & 72.52/76.02 &       & 23.55/22.98 & 50.99/47.77 &       & 17.69/17.52 & 44.03/40.33 &  \\
				PESA& 37.15/43.46 & 59.76/65.45 &       & 45.03/56.80 & 71.95/75.02 &       & 22.67/21.81 & 50.02/46.01 &       & 18.00/16.95 & 44.10/38.67 &  \\
				MSA-Net& 37.40/40.22 & 60.16/63.67 &       & 48.45/48.75 & 73.23/72.24 &       & 18.51/20.11 & 45.74/45.57 &       & 15.26/16.17 & 41.00/39.18 &  \\
				\textbf{LLHA-Net}  & \textbf{43.16}/\textbf{45.16} & \textbf{65.39}/\textbf{67.24} &       & \textbf{50.88}/\textbf{57.05} & \textbf{74.99}/\textbf{76.42} &       & \textbf{24.09}/\textbf{23.55} & \textbf{51.74}/\textbf{48.03} &       & \textbf{18.93}/\textbf{18.09} & \textbf{44.46}/\textbf{40.36} &  \\
				\bottomrule    
			\end{tabular}%
		}
		\label{tab:camera pose}%
	\end{table*}%
	Camera pose estimation is another important task in our experiments, which involves determining the camera pose by matching feature points in images. 
	In this task, we use the probability set obtained from the previous stage of outlier removal as weights in the weighted eight-point algorithm to compute the essential matrix and camera pose.\par
	Table \ref{tab:camera pose} presents the mAP performance at error thresholds 5° and 20° after RANSAC post-processing. Without RANSAC, LLHA-Net achieves an average accuracy of 74.99\% and 44.46\% in indoor and outdoor unknown scenes respectively, with the error threshold 20°. 
	These results are 6.15\% and 3.44\% higher than those achieved by the representative OA-Net. 
	LLHA-Net exhibits better improvements compared to the baselines on most data, particularly in known scenes or without RANSAC.
	When comparing the experimental results with and without RANSAC, we observe that the improvement in our experimental results is smaller compared to other experiments. 
	In YFCC100M, with RANSAC post-processing, our results show improvements of 2.00\%, 1.85\%, 6.17\%, and 1.43\% respectively over those without post-processing. This improvement is much lower than some competing methods, such as PESA with improvements of 6.31\%, 5.69\%, 11.77\%, and 3.07\%. 
	It indicates that LLHA-Net is not heavily reliant on post-processing, which indicates that we can obtain higher-quality correspondences during the outlier removal process, effectively removing a significant number of feature correspondences with noticeable errors.\par
	In summary, the outlier removal performed by LLHA-Net lays a solid foundation for camera pose estimation and provides higher-quality feature correspondences. This contributes to achieving the best results among all the competing methods.
	
	\begin{figure}[!t]
		\vspace{-4ex}
		\centering
		\includegraphics[width=0.9\linewidth]{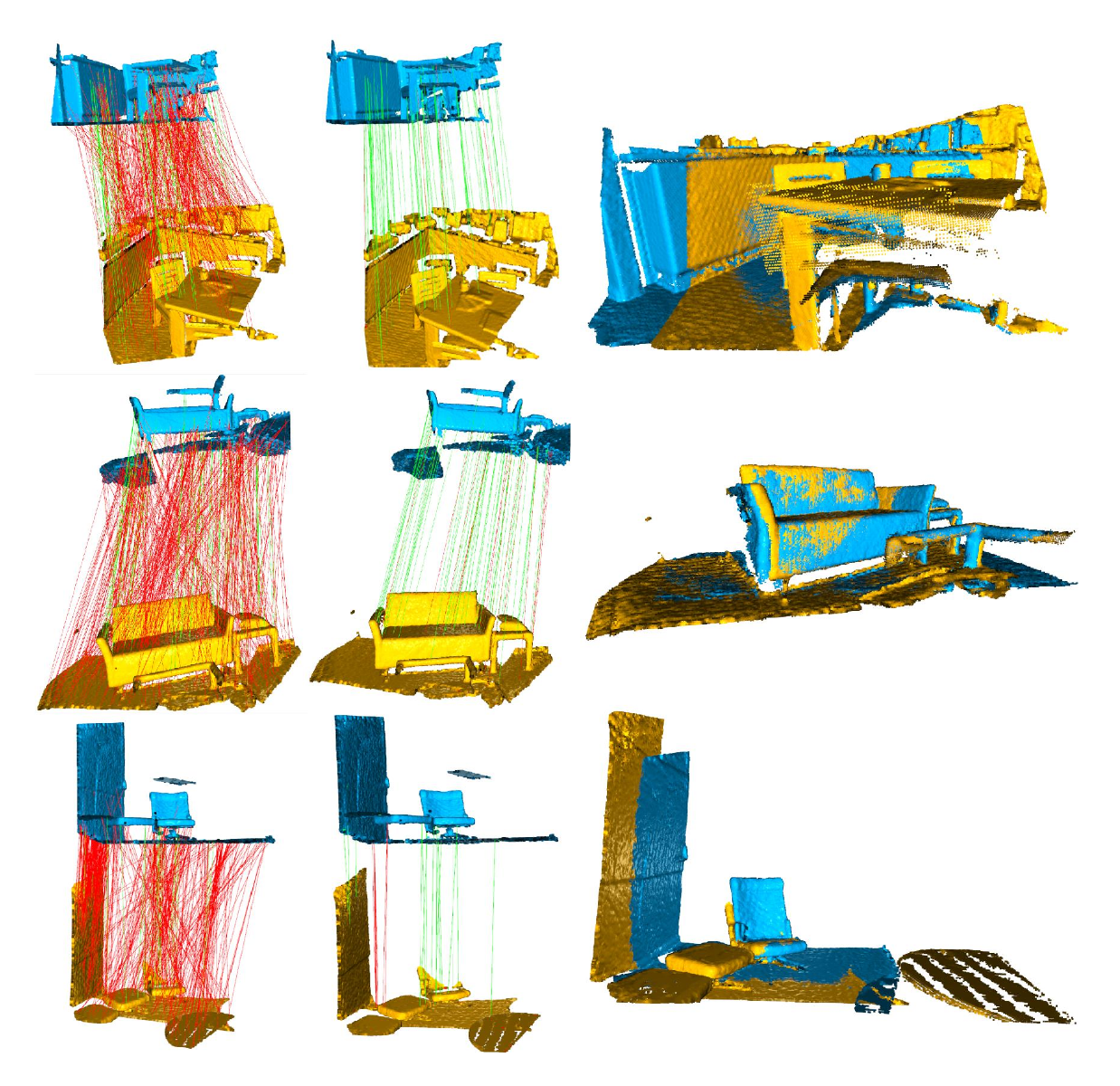}
		\caption{Visualization of 3D point cloud registration and fusion. From left to right, the images depict the initial matching, outlier removal by the proposed method, and the final 3D fusion.}
		\label{fig:3d}
		\vspace{-2ex}
	\end{figure}
	
	\begin{figure}[!t]
		\vspace{-4ex}
		\centering
		\includegraphics[width=0.9\linewidth]{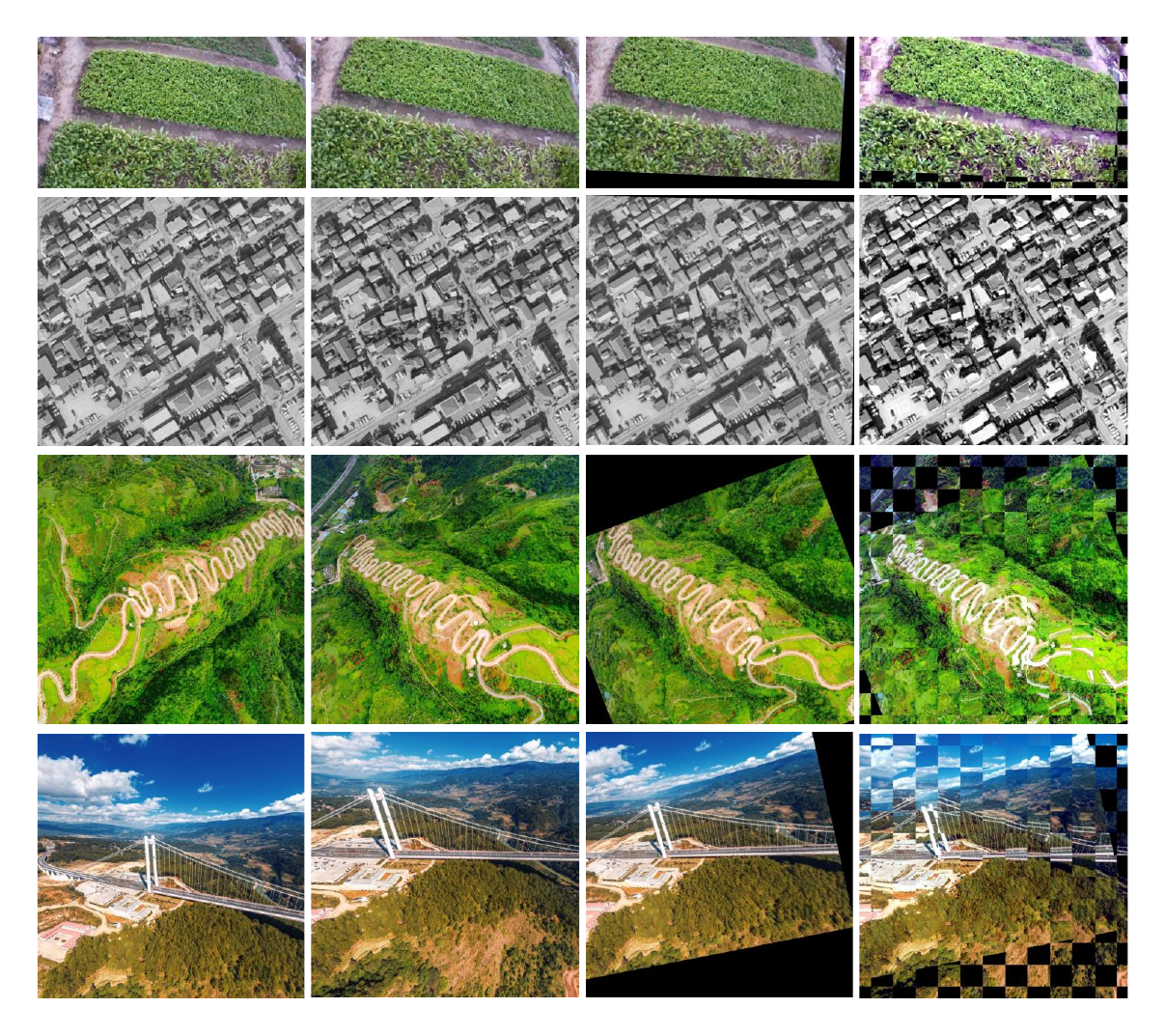}
		\caption{Visualization of remote sensing image registration. Each row represents a scene, where the images in the first and second columns are the source and target of the registration. The third column shows the affine transformation results, and the fourth column presents a checkerboard overlay of the registration results and the target.}
		\label{fig:yaogan}
		\vspace{-2ex}
	\end{figure}
	
	\subsection{Applications}
	
	To demonstrate the generalization potential of our approach in 3D scenes, we perform an experimental evaluation of the 3DMatch dataset \cite{3dmatch}. Using the PointDSC framework \cite{pointdsc}, our model was first trained on the training set of 3DMatch and then tested on the test set. The test set contains 1623 pairs of partially overlapping point cloud fragments from eight diverse scenes, each annotated with detailed ground truth labels. In the three different scenes shown in Figure \ref{fig:3d}, our method effectively removes most of the misalignment points. By demonstrating the visualization of representative scenes, our approach not only efficiently removes misalignment points, but also achieves superior 3D fusion. Our approach enables seamless splicing of point cloud fragments, providing solid technical support for creating more complete and accurate 3D scenes.

	For the image registration task, we conduct experiments on a remote sensing dataset \cite{remoteLMR}. These scenes include, but are not limited to, urban landscapes, natural landforms, and complex agricultural areas, which are often considered challenging in image registration due to the amount of detail and texture variation they contain.
	Figure \ref{fig:yaogan} shows remote sensing registration of relevant scenarios.
	LLHA-Net can obtain high quality remote sensing registration results, showing its strong adaptability and robustness.
	
	\subsection{Ablation Studies}
	
	To further understand the contributions of each component within LLHA-Net, we conducted extensive the ablation studies. These experiments were designed to isolate and evaluate the effects of key modules, including the PIHA modules, the LLF module, and the hierarchical attention mechanism, on the overall performance of the model. Specifically, we systematically removed or modified individual components of LLHA-Net. By comparing the performance metrics of these modified models with those of the complete architecture, we quantified the impact of each module on accuracy and robustness.

	\subsubsection{Module Test}
	
	To evaluate the effectiveness of our designed modules, we conducted experiments using different combinations of them. 
	Table \ref{tab:ablation1} showcases the results of camera pose estimation on the YFCC100M dataset, with the error threshold 5°, using PointCN as the baseline model. 
	The introduced modules include DP\&OA\&UDP, iteration network, LLF, HA, and PIHA. 
	The results demonstrate that the performance on both known and unknown scenes increased from 21.79\% to 42.90\% and from 30.08\% to 50.63\%, respectively. Specifically, the LLF, HA, and PIHA modules contributed to an improvement in the experimental results by (2.60\%, 1.27\%), (2.23\%, 3.69\%), and (1.91\%, 4.79\%), respectively.
	LLF generates and fuses feature information at the channel level, ensuring that crucial information at each stage is effectively maintained and enhanced.
	HA extracts both global perceptual and local semantic information across various levels. This dual extraction not only guarantees the comprehensiveness of the overall information but also meticulously analyzes the matching pairs of distinct feature points.
	PIHA is processed in a permutation invariant form, allowing for the robust handling of feature permutations while maintaining the integrity of the analysis.
	The results of our ablation studies provide significant insights into the contributions of each module to the performance of LLHA-Net. Specifically, the LLF module enhances feature extraction at the channel level, effectively preserving and amplifying critical information across the processing stages. Meanwhile, the HA module captures both global perceptual and local semantic information, enabling a comprehensive analysis of feature correspondences. Additionally, the PIHA module ensures robust handling of feature permutations, which is essential for maintaining analytical integrity under varying conditions.
	Overall, our experimental results clearly demonstrate the effectiveness of the designed modules in enhancing the accuracy of camera pose estimation. The combination of these modules resulted in significant performance improvements on both known and unknown scenes when compared to the baseline PointCN model.
	\begin{table}[!t]
		\centering
		\caption{Ablation studies on different module combinations. P\&O\&U: using the DP\&OA\&UDP module. Iter: using the iterative network. LLF: using LLF module. HA: using HA module. PIHA: using PIHA module.}
		\resizebox{0.95\linewidth}{!}{
			\begin{tabular}{cccccccc}
				\toprule
				PointCN &P\&O\&U & Iter &LLF &HA &PIHA  & Known & Unknown\\
				\midrule
				\checkmark     &       &       &       &       &       & 21.79 & 30.08 \\
				\midrule
				\checkmark       & \checkmark       &       &       &       &       & 31.99 & 36.95 \\
				\midrule
				\checkmark       & \checkmark      & \checkmark       &       &       &       & 36.16 & 40.88 \\
				\midrule
				& \checkmark       & \checkmark       & \checkmark       &       &       & 38.76 & 42.15 \\
				\midrule
				& \checkmark       & \checkmark       & \checkmark       & \checkmark       &       & 40.99 & 45.84 \\
				\midrule
				& \checkmark       & \checkmark       & \checkmark       &       & \checkmark       & \textbf{42.90}  & \textbf{50.63} \\
				\bottomrule
			\end{tabular}%
		}
		\label{tab:ablation1}%
	\end{table}%
	
	\subsubsection{LLHA-Net vs T-Net and PIHA vs PCSE}
	\begin{figure}[htbp]
		\centering
		\includegraphics[width=0.9\linewidth]{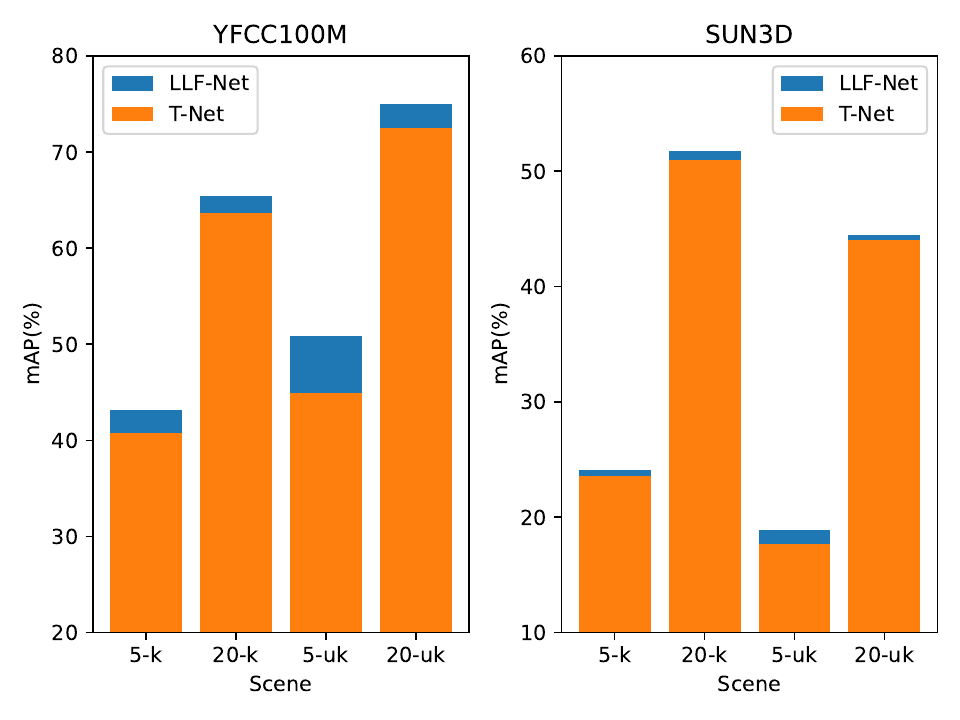}
		\vspace{-2ex}
		\caption{Camera pose estimation without RANSAC comparison between LLHA-Net and T-Net. 5-k denotes known scenes at the error threshold 5°, while ``others" indicates a similar comparison.}
		\label{fig:tvsl}
	\end{figure}
	LLHA-Net is an extension of T-Net, with significant improvements and enhancements. We have introduced two architectures that focus on different aspects of feature extraction and integration, enabling us to extract feature information more effectively from the data.
	In particular, our designed PIHA module surpasses the PCSE module in T-Net. 
	The PIHA module retains the permutation invariance and feature extraction characteristics of the PCSE module, while introducing advantages such as hierarchical fusion and layer-by-layer fusion. 
	These enhancements have greatly improved the performance of our model.
	To demonstrate the effectiveness of our approach, we have selected a subset of the camera pose estimation data and visualized it in Figure \ref{fig:tvsl}. 
	For more detailed data, please refer to Tables \ref{tab:outlier removal} and \ref{tab:camera pose}. Comparing the results with T-Net, we can observe that LLHA-Net outperforms it in all data indicators.
	PCSE lacks distinguishing treatment of feature information, simply generating all feature information through a convolutional layer, and is unable to extract deeper information.
	In contrast, the PIHA module ensures robust handling of feature permutations, which is essential for preserving the integrity of the analysis under varying conditions.
	Overall, our model architecture and PIHA module offer a deeper and more hierarchical approach to data feature extraction and model performance improvement.
	
	\subsubsection{Impact of LLF Module}
	In the design of the HA module, we have replaced PointCN with LLF. To ensure scientific validity, we conducted four experiments: PointCN+PointCN, LLF+PointCN, PointCN+LLF, and LLF+LLF, at both ends of the HA module. Table \ref{tab:Impact of LLF module} presents detailed data on the known scenes of YFCC100M. 
	In all experiments, LLF+LLF achieved the highest scores for all metrics except recall. Only PointCN+LLF achieved the highest recall score of 90.56\%, while LLF+LLF obtained the second-highest score of 90.14\% with a marginal difference. 
	Despite the 3.4\% increase in parameter quantity, LLF+LLF showed significant overall improvement in all indicators compared to PointCN+PointCN and LLF+PointCN, which is undoubtedly valuable.
	LLF is more comprehensive than PointCN in extracting and generating information on channels. On the other hand, PointCN solely processes information through simple convolution and normalization, and is unable to extract deeper feature information.
	LLF performs better than PointCN in terms of information extraction and fusion, and the layer-by-layer fusion mechanism helps reduce the loss of semantic information.
	Based on the above experiments, the LLF module is shown to enhance feature extraction at the channel level, effectively preserving and amplifying key information throughout the entire processing stage. Meanwhile, the HA module captures both global perceptual and local semantic information, enabling a comprehensive analysis of feature correspondences.
	\begin{table}[!t]
		\centering
		\caption{Impact of LLF module.}
		\resizebox{0.95\linewidth}{!}{
			\begin{tabular}{cccccccc}
				\toprule
				\multirow{2}[1]{*}{Method} & \multirow{2}[1]{*}{Parameters} & \multirow{2}[1]{*}{P(\%)} & \multirow{2}[1]{*}{R(\%)} & \multirow{2}[1]{*}{F(\%)} & \multicolumn{3}{c}{mAP(\%)} \\
				\cmidrule{6-8}          &     &  &       &       & 5°    & 10°   & 20° \\
				\midrule
				PointCN+PointCN &3.81M &  62.23  &  89.97     &73.57  & 43.10  & 53.77  &65.08  \\
				LLF+PointCN &3.87M &  62.90  & 89.89  & 74.01 & 42.43   & 53.85   &65.35  \\
				PointCN+LLF &3.87M &  62.28   & \textbf{90.56}   & 73.80 & 42.99  & 53.86 &65.21  \\
				LLF+LLF &3.94M &\textbf{63.24} & 90.14 & \textbf{74.33}& \textbf{43.16} & \textbf{53.95}&  \textbf{65.39}\\
				\bottomrule
			\end{tabular}%
		}
		\label{tab:Impact of LLF module}%
	\end{table}%

	\subsubsection{Impact of Global Average Pooling}
	To evaluate the impact of global average pooling (GAP) in the proposed HA module, we conducted the comparative experiments with global maximum pooling (GMP). Tables \ref{tab:outlier_removal2} and \ref{tab:camera_pose2} show the quantitative results obtained by GAP and GMP. The experimental results indicate that GMP generally performs worse than GAP. This is because the input data contains a high proportion of outliers, and GMP tends to select these outliers, leading to suboptimal results. On the other hand, GAP provides a stable and more representative summary of the global perceptual information. By averaging the feature values, it mitigates the impact of outliers and generates more reliable features for subsequent processing. Overall, utilizing GAP in the HA module to extract global perception information can enhance the robustness and accuracy of the model.
	In summary, the ablation studies highlight the critical contributions of each component within LLHA-Net. The combined functionality of the LLF, HA, and PIHA modules results in significant improvements in both accuracy and robustness, positioning LLHA-Net as a competitive solution for feature correspondence tasks. Furthermore, these findings not only validate the effectiveness of our architectural design but also identify promising directions for future research and optimization.

	\begin{table}[!t]
		\centering
		\caption{Comparative results of outlier removal on the YFCC100M dataset.}
		\resizebox{0.95\linewidth}{!}{	
			\begin{tabular}{ccccccccc}
				\toprule
				Datasets & \multicolumn{8}{c}{YFCC100M}\\
				\midrule
				\multirow{2}[4]{*}{Matcher} & \multicolumn{3}{c}{Known Scene} &       & \multicolumn{3}{c}{Unknown Scene} &  \\
				\cmidrule{2-4}\cmidrule{6-8}         & P(\%) & R(\%) & F(\%) &       & P(\%) & R(\%) & F(\%) &\\
				\midrule
				GMP  & 62.67 & 90.12 & 73.93 &       & 58.32 & 87.01 & 69.83 \\
				GAP  & \textbf{63.24} & \textbf{90.14} & \textbf{74.33} &       & \textbf{59.28} & \textbf{87.34} & \textbf{70.62} \\
				\bottomrule
			\end{tabular}%
		}
		\label{tab:outlier_removal2}%
	\end{table}%

	\begin{table}[!t]
		\centering
		\caption{Performance comparison for camera pose estimation on YFCC100M datasets. The mAP performance at error thresholds 5° and 20° are reported \textbf{with/without} RANSAC post-processing.}
		\resizebox{0.95\linewidth}{!}{
			\begin{tabular}{ccccccc}
				\toprule
				Datasets & \multicolumn{6}{c}{YFCC100M}\\
				\midrule
				\multirow{2}[4]{*}{Matcher} & \multicolumn{2}{c}{Known Scene} &       & \multicolumn{2}{c}{Unknown Scene} & \\
				\cmidrule{2-3}\cmidrule{5-6} & 5°    & 20°   &       & 5°    & 20°   & \\
				\midrule
				GMP  & 42.71/44.35 & 63.44/65.69 &       & 49.18/55.55 & 74.24/76.20 &\\
				GAP  & \textbf{43.16/45.16} & \textbf{65.39/67.24} &       & \textbf{50.88/57.05} & \textbf{74.99/76.42} &\\
				\bottomrule    
			\end{tabular}%
		}
		\label{tab:camera_pose2}%
	\end{table}%

	\section{Conclusion}
	\label{conclusion}
	
	In this paper, we propose a novel LLHA-Net designed to enhance the representation ability and matching accuracy of feature points. Particularly, LLHA-Net comprises three core components: LLF, HA and PIHA. The proposed LLF module extracts information from each stage while retaining previous information for channel fusion; the HA module leverages the attention mechanism to integrate information from different levels, facilitating effective information fusion; the PIHA module combines the strengths of each module and incorporates point-by-point convolution to maintain permutation invariance. 
	The experimental results demonstrate that LLHA-Net exhibits superior feature representation ability, enabling it to capture richer semantic information and establish high-quality feature point correspondences. 
	It provides a new method for image registration and can be used as a component in related fields such as 3D registration, motion reconstruction to improve the effectiveness of registration.
	However, since our overall architecture is integrated and constructed stage by stage, the parameter reduction is relatively small. While LLHA-Net improves matching accuracy and robustness, the increase in parameters and computational cost may limit its suitability for real-time or resource-constrained applications. In addition, the feature fusion strategy may occasionally misclassify some inliers as outliers.
	In future work, we aim to optimize the network structure and loss function, with particular emphasis on balancing the approach and extent of information fusion to enhance the recall rate in outlier removal while improving computational efficiency.


	\section*{Declaration of competing interest}
	The authors declare that they have no known competing financial interests or personal relationships that could have appeared to influence the work reported in the paper.
	
	\section*{Data availability}
	We use publicly available dataset.
	
	\section*{Acknowledgements}
	This work was supported in part by the National Natural Science Foundation of China (Grant Nos. U22A2095, 62476112, 62272200, 62202249, 62472312), in part by the Guangdong Basic and Applied Basic Research Foundation (Grant Nos. 2024A1515011740, 2025A151501018), and in part by the Fundamental Research Funds for the Central Universities (Nos. 21624404, 23JNSYS01), in part by the Guangdong Key Laboratory of Data Security and Privacy Preserving (Grant No. 2023B1212060036), and in part by Guangdong-Hong Kong Joint Laboratory for Data Security and Privacy Preserving (Grant No. 2023B1212120007).

	\bibliography{References}
	
\end{document}